\documentclass[preprint,10pt]{elsarticle}

\usepackage{graphicx}
\usepackage{booktabs}
\usepackage{subcaption}
\usepackage{hyperref}
\usepackage{float}
\usepackage{todonotes}
\usepackage{amsmath}
\usepackage{amssymb}
\usepackage{dsfont}
\usepackage{cleveref}
\usepackage[table,dvipsnames]{xcolor}
\usepackage[linesnumbered, ruled, vlined]{algorithm2e}
\DeclareGraphicsExtensions{.eps}  
\graphicspath{{media/}}
\usepackage[utf8]{inputenc}
\usepackage{newunicodechar}

\usepackage{orcidlink}
\newunicodechar{τ}{\ensuremath{\tau}}

\begin{document}

\begin{frontmatter}

\title{Faithful, Sufficient and Understandable: Rethinking Graph Counterfactual Explanations via Discrete Diffusion Inversion}

\author{David Bechtoldt\textsuperscript{a,b}~\orcidlink{0009-0002-0305-0778}}
\author{Sidney Bender\textsuperscript{a,b}~\orcidlink{0009-0007-1727-8577}}
\address{\textsuperscript{a}Machine Learning Group, TU Berlin, 10587 Berlin, Germany}
\address{\textsuperscript{b}BIFOLD–Berlin Institute for the Foundations of Learning and Data, Berlin, Germany}

\begin{abstract}
Graph Neural Networks (GNNs) achieve strong predictive performance on graph-structured data across domains such as chemistry, biology, and network analysis, yet they provide no intrinsic explanation of their predictions. This limits their adoption in high-stakes and safety-critical settings.
Counterfactual explanations address this by revealing the minimal structural modifications that would change a model's prediction. On graphs, however, such a modification is hard to produce.
The search space is discrete and combinatorial, and a valid answer must respect categorical node and edge types together with domain rules such as chemical valency in the case of molecular graphs. Existing explainers give up one of two things. Either edits are not held on the data manifold, or the search does not span the full edit space. We propose Graph Diffusion Counterfactual Explanation via Inversion (GDCE-I), which gives up neither. A
discrete denoising diffusion model with a novel discrete inversion scheme enables distribution-aware edits leveraging the whole domain edit space. We further address the incomplete and inconsistent evaluation of graph counterfactuals by deriving a framework of explanation desiderata and applying it to every method under one shared protocol. Across four benchmarks, GDCE-I outperforms related work by a large margin on the defined framework. For the molecular domain, we further qualitatively show that GDCE-I attains interpretable in-distribution solutions. 
\end{abstract}


\end{frontmatter}
\newpage

\section{Introduction}
Graph Neural Networks \cite{GNN} have emerged as the de facto standard for learning graph-structured data across domains such as chemistry, biology, and network analysis, achieving breakthroughs in tasks such as molecular property prediction, drug discovery, and toxicity analysis \cite{Schnet, Drugdiscovery, MoleculeNet}. Despite their predictive abilities, GNNs operate predominantly as black-box models. This lack of transparency undermines the decision-making process, severely limiting its adoption in high-stakes scenarios where interpretability and safety are paramount \cite{GraphXAI, GraphXAI01, GraphXAI02}. For example, the discovery of a new structural class of antibiotics was guided not by the prediction of twelve million candidates by a graph neural network, but by the chemical substructures the model had associated with selective activity \cite{XAIDiscovery}. 

To uncover such explanations, Explainable AI (XAI) offers a diverse set of approaches to explain a neural network. A prominent branch of XAI is attribution-based methods, such as Layer-wise Relevance Propagation \cite{LRP} that was further developed for GNNs (GNN-LRP) \cite{LRPGraph}. These techniques answer the "why" behind a model's decision by identifying and highlighting the most influential features or structural motifs.

\begin{figure}[H]
    \centering
    \includegraphics[width=\textwidth]{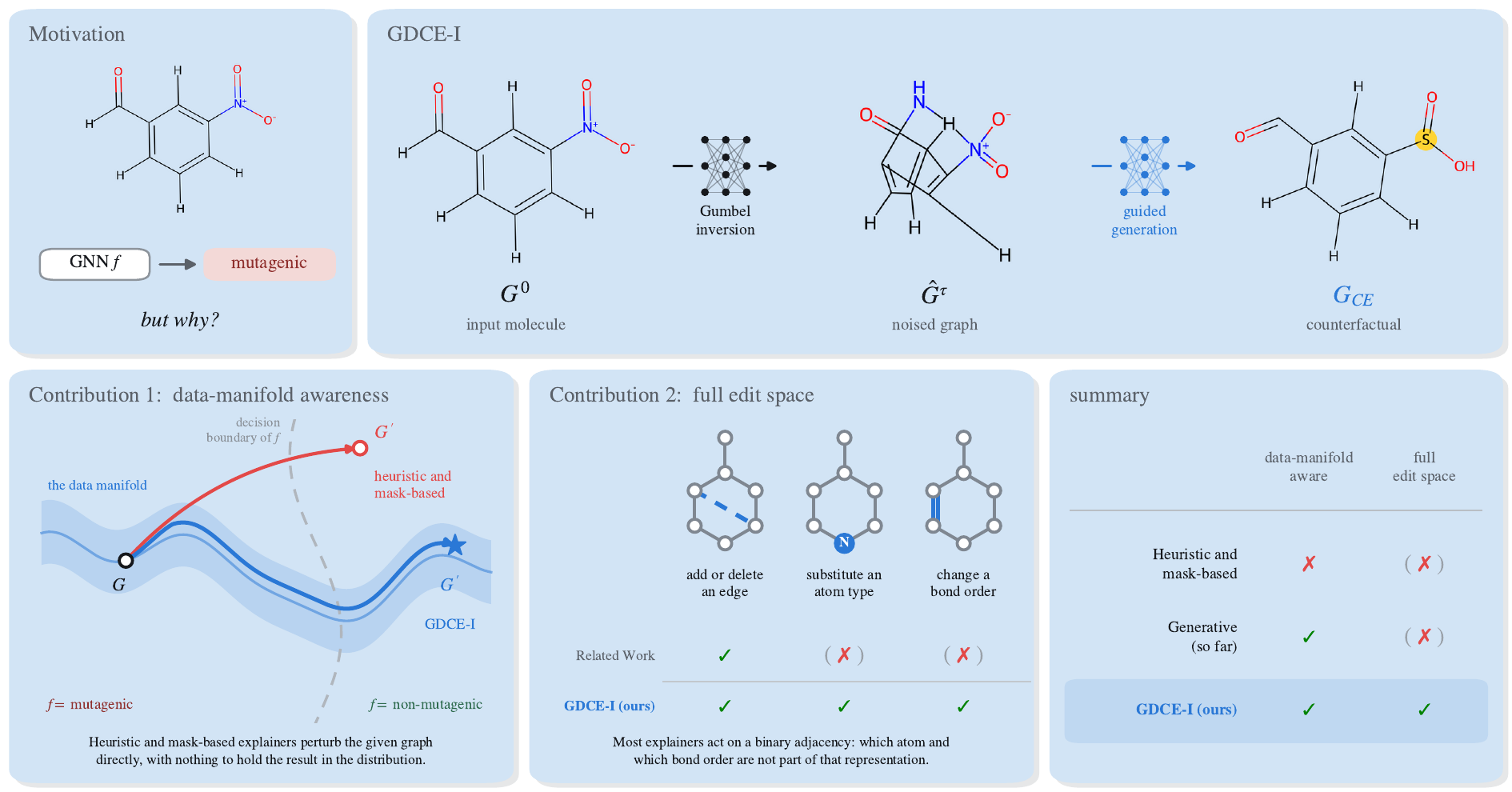}
    \caption{GDCE-I generates graph counterfactuals by inverting a discrete
diffusion trajectory. Respecting the data manifold and covering the full edit
space is what makes the explanations domain-valid.}
    \label{fig:figure1}
\end{figure}

Another approach to explaining a model is counterfactual explanations \cite{Counterfactuals, Diffeomorphic, Survay_Graph_Counterfactual}. They provide a realistic counterexample for a given input, which differs as little as possible to alter the classifier's prediction.
Rather than attributing relevance to existing structures, counterfactuals answer: "What minimal changes to the input graph would alter the model's prediction?" In computational chemistry, this could translate to targeted structural edits that can neutralize a toxicophore.

However, translating counterfactual generation to the graph domain introduces unique challenges that are largely absent in continuous domains such as computer vision. Graphs are inherently discrete, non-euclidean, and combinatorial objects  \cite{Survay_Graph_Counterfactual, Graph_Counterfactuals}. Crucially, real-world graph representations, such as molecules or proteins, reside in a highly sparse topological space \cite{DiGress}. Consequently, even minor topological perturbations can easily violate strict domain-specific rules, such as chemical valency, resulting in structurally invalid and semantically meaningless instances.

Early approaches relied on heuristic perturbation strategies restricted solely to edge deletions \cite{Survay_Graph_Counterfactual, CF-GNNExplainer}. This constrains the counterfactual search space, prohibiting the addition of necessary structural motifs or the alteration of edge and node types.
To improve the realism of the explanations, a parallel line of work turns to generative models that keep counterfactuals close to the data distribution \cite{Graph_Counterfactuals, D4Explainer, CLEAR}. Although more recent heuristic methods support edge additions and node-type changes \cite{C2Explainer}, most methods in both families still operate on a binary adjacency matrix that cannot represent categorical edge types.
To systematically address these limitations, we propose \textit{Graph Diffusion Counterfactual Explanation via Inversion} (GDCE-I) \footnote{A preliminary version of this work was presented at the ESANN 2026 Conference \cite{GDCE}. This paper presents the fully extended framework.}. By combining discrete denoising diffusion models \cite{DiGress} with classifier-free guidance \cite{FreeGress, Classefier-Free} and an edit-friendly inversion technique, our approach allows for comprehensive modifications, including targeted edge additions, deletions, edge and node type alterations, within a data-manifold-aware generation process (Figure~\ref{fig:figure1}).

\begin{figure}[H]
    \centering
    \includegraphics[width=\textwidth]{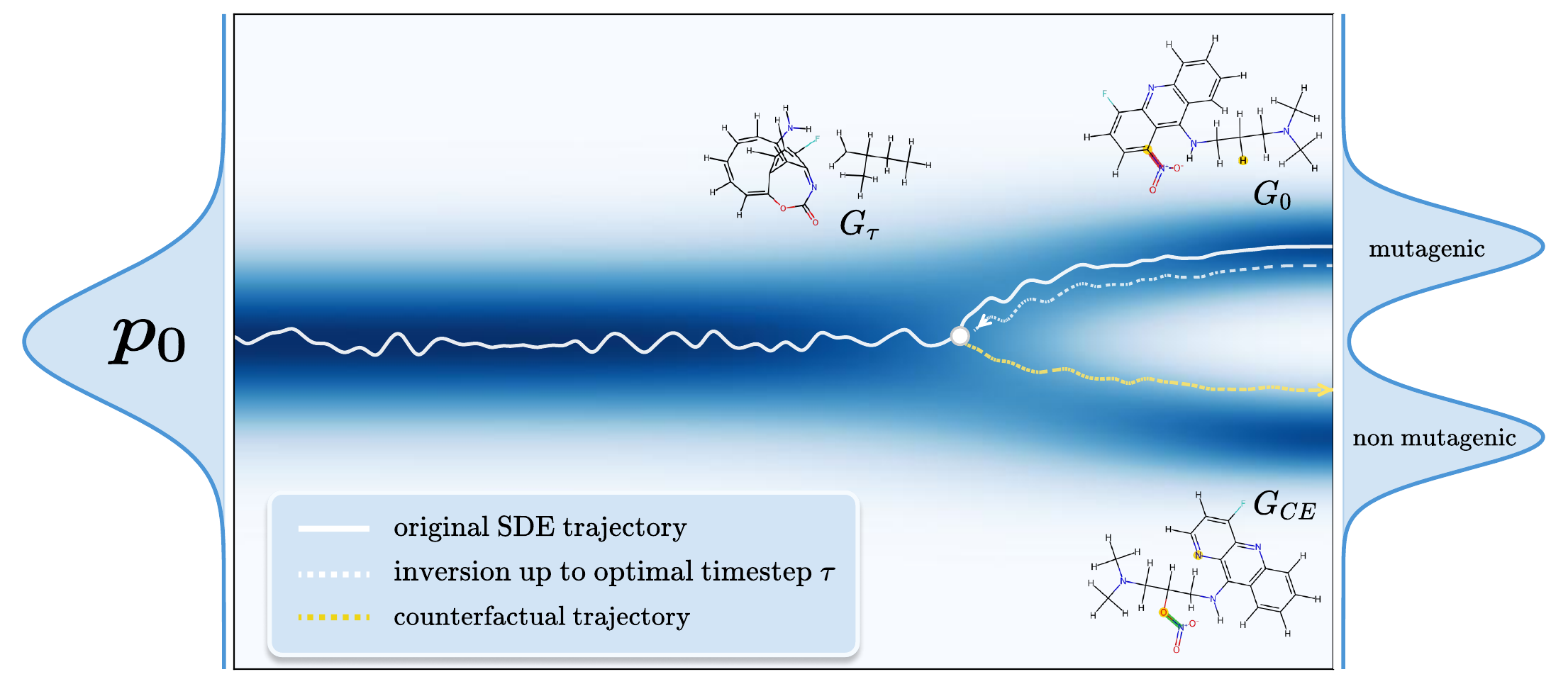}
    \caption{High-level illustration of GDCE-I. By manipulating the latent space of the learned diffusion model, we can generate distribution-aware and sparse counterfactuals.}
    \label{fig:figure2}
\end{figure}

GDCE-I integrates discrete diffusion models with classifier-free guidance, so that the model captures the conditional distribution of valid graphs directly. Its core contribution is an \textit{edit-friendly} inversion scheme for discrete diffusion, an inversion that records the sampling stochasticity of a given graph, so the graph can be reconstructed and then minimally edited by changing only the target condition. We derive it by exploiting the Gumbel-Max trick in a discrete diffusion setting. We traverse a reference trajectory of the given graph in the reverse direction and, at every step, construct the Gumbel noise that forces the guided reverse posterior to reproduce that trajectory, using a truncated-Gumbel ($\mathrm{A}^\star$-sampling) construction \cite{AStarSampling}. The resulting posterior noise space reconstructs the original graph exactly under its own condition. Re-injecting this same noise during a reverse generation pass guided toward a new target property therefore preserves the structural identity of the original graph as far as the new target permits, favoring edits that are both minimal and in distribution. . The trade-off between sparsity and the desired property flip is controlled via a dynamic skip-parameter~$\tau$. A high-level illustration can be contemplated in Figure~\ref{fig:figure2}. Our main contributions are summarized as follows:

\begin{itemize}
    \item A Novel Generative Inversion Framework: We introduce GDCE-I, the first method to unify discrete graph diffusion, classifier-free guidance, and Gumbel-Max inversion. This framework enables precise, edit-friendly manipulation of discrete graph structures.
    \item Benchmarking on Classifier-Distilled Datasets: Through a rigorous evaluation of the Mutagenicity \cite{mutagenicity}, Benzene \cite{BENZENE, BENZENXAI}, PROTEINS \cite{PROTEINS, PROTEINS01} and TWITTER \cite{TWITTER, TWITTER01} dataset against baselines such as $CF^2$ \cite{CF2}, C2Explainer \cite{C2Explainer}, XPlore \cite{XPlore}, UCExplainer \cite{UCExplainer}, and D4Explainer \cite{D4Explainer}, we show that GDCE-I is the only method that jointly produces faithful and understandable counterfactuals.  We further provide qualitative evidence that GDCE-I recognizes and alters functional groups via realistic chemical substitutions for Mutagenicity and Benzene.
\end{itemize}

\section{Related Work}
\label{sec:related}

\paragraph{Counterfactual Explanations}

Although there are counterfactual explanation methods for domains such as natural language~\cite{dehghanighobadi2025can} and proteins~\cite{klos2026protein}, the most established fields are tabular data and computer vision. Tabular data often is low-dimensional and specific properties of the input features are known, so there exist a lot of approaches directly using this knowledge, e.g., in the form of Integer Linear Programming~\cite{russell2019efficient} or other explicit closed-form optimization schemes such as in DiCE~\cite{mothilal2020explaining}.
For both tabular data and computer vision, there are various approaches that use a generative model to filter gradients to stay in the data manifold while moving towards the desired class~\cite{Diffeomorphic, rodriguez2021beyond, augustin2022diffusion, jeanneret2022diffusion, weng2024fast, bender2025desiderata, zeid2026sce, jeanneret2023adversarial}.
These approaches cannot be easily transferred to graphs due to the discrete nature. In addition, there are approaches for both tabular and computer vision in the form of distillation~\cite{jeanneret2024text, bender2026visual}.
These approaches are in principal transferable to graphs, but one has to choose a suitable generative model.

\paragraph{Heuristic and mask-based methods for Graphs}
Early approaches rely on heuristic perturbations. CF-GNNExplainer and CF$^2$ \cite{CF-GNNExplainer, CF2} learn a continuous mask over the existing edge set, which restricts the search space to edge deletions only and, therefore, cannot introduce the structural motifs that a valid counterfactual may require. C2Explainer \cite{C2Explainer} broadens this space to edge additions and node-feature perturbations, and XPlore \cite{XPlore} pursues the same direction, driving gradient-guided perturbations of the adjacency and the node-feature matrix jointly. InduCE \cite{InduCE} instead attacks the per-instance optimization itself, learning a reinforcement-learning policy over edge additions and deletions that transfers to unseen inputs without instance-specific training. It targets node rather than graph classification and is therefore outside the scope of our comparison. All of these methods operate on a binary adjacency matrix and thus cannot predict categorical edge types, i.e.\ they cannot distinguish a single, double, or aromatic bond between the same pair of atoms, a distinction that is essential for chemical validity. UCExplainer \cite{UCExplainer} is the exception among the mask-based methods. It perturbs node features, topology, and bond types alike and is therefore, together with GDCE-I, the only method that can be evaluated against an edge-feature-aware classifier.

A complementary line explains the model rather than the instance. Global graph counterfactual explanation \cite{GlobalCFE} examines a compact set of subgraph substitution rules that flip the prediction across many graphs at once, trading per-instance faithfulness for coverage. We target instance-level counterfactuals throughout.

\paragraph{Generative methods for Graphs}
A second line of work generates counterfactuals from a learned model of the data distribution. CLEAR \cite{CLEAR} employs a graph variational autoencoder (VAE) that decodes a Bernoulli adjacency matrix together with node features but does not assign bond types. CGCF \cite{Graph_Counterfactuals} also traverses a continuous VAE latent space and, unlike CLEAR, it decodes categorical edge types, but it shares the more fundamental drawback that a continuous latent relaxation is a questionable model for the highly constrained, discrete nature of graphs. RSGG-CE \cite{RSGGCE} instead adopts a GAN, again without categorical edge types. In the molecular domain, MEG \cite{MEG} employs a reinforcement-learning agent that edits the molecular graph with valence-valid operations. A related line of work for molecules operates outside the graph representation altogether, namely MMACE, which enumerates chemically valid neighbors through SELFIES string mutations. LLM-GCE \cite{LLM-GCE} also operates on textual input, generating counterfactuals with large language models for Molecules. Diffusion-based methods are the most directly related. D4Explainer \cite{D4Explainer} trains a discrete denoising diffusion model with a counterfactual loss term to generate in-distribution explanations. However, it operates on a binary adjacency matrix without categorical nodes or edge types, and its counterfactual objective is incorporated only at training time.

\paragraph{Evaluating Explanations} A major challenge in designing explanation techniques is the issue of evaluation.
Ground-truth explanations are rarely available, and the notion of a good explanation depends on the use case.
The question of evaluation has been central, with early foundations including \cite{shapley:book1952} and \cite{Swartout1993}, with the latter proposing a set of practical desiderata for explanation techniques.
While the evaluation of explanations has been extensively covered for attribution methods (e.g.\ \cite{DBLP:journals/tnn/SamekBMLM17,DBLP:journals/csur/NautaTPNPSSKS23}), for visual counterfactuals in ~\cite{bender2025desiderata}, and counterfactuals for tabular data (e.g. {\cite{mothilal2020explaining}}), there have been comparatively fewer such studies in the context of graph counterfactual explanations.

In summary, while similar metrics to the one suggested by us have been proposed in the context of tabular and vision data, previous work on graph counterfactual evaluation has predominantly focused on a form of \textit{flip-ratio} and \textit{sparsity}.

\section{Desiderata of Counterfactuals}
\label{section:desiderata}

To enhance the usefulness of graph counterfactual explainers (GCEs), and analogous in large parts to the desiderata-driven framework established for visual counterfactuals by Bender et al.~\cite{bender2025desiderata}, we take as a starting point the holistic explanation desiderata formulated by Swartout and Moore (1993) \cite{Swartout1993}. These desiderata, originally proposed in the context of explaining expert systems, are `fidelity', `understandability', `sufficiency', `low construction overhead', and `efficiency'. In the following, we contribute an instantiation of the first three desiderata specifically tailored to counterfactual explanations on graphs and operationalize them using concrete metrics: Non-Adversarial Rate (NA), a threefold Sparsity breakdown (Edge, Node, Edge Type), and Non-Adversarial Flip Rate (NAFR).

\subsection{Fidelity}
\label{section:fidelity}

For an explanation to be useful, it should faithfully describe what the model does. As noted in \cite{Swartout1993}, an incorrect or misleading explanation is worse than no explanation at all. Instantiating this desideratum to counterfactual explanations following \cite{bender2025desiderata}, we propose the following formalization: Assuming a factual graph $G$ and a classifier $f$ that varies locally following some dominant direction $w$, the counterfactual graph $G'$ is faithful, i.e., represents what the model does, if it meets the following three criteria:
\begin{enumerate}
\item The counterfactual should be plausible and thus be on the data manifold, i.e., $G' \in \mathcal{M}$, adhering to structural and domain-specific rules (such as valency).
\item The counterfactual's immediate neighborhood, i.e. $\{\xi \in \mathcal{M}: d(\xi, G') < \delta\}$, should also be dominantly counterfactual.
\end{enumerate}
In practice, testing these properties directly is not always possible due to discrete, combinatorial graph spaces and limited knowledge of $\mathcal{M}$. 

It is interesting to note that most state-of-the-art GCEs, such as CLEAR \cite{CLEAR} and D4Explainer \cite{D4Explainer}, ensure (i) by limiting the search for $G'$ to the data manifold $\mathcal{M}$ through a generative model or not all \cite{CF-GNNExplainer, C2Explainer,  UCExplainer}. However, they do not address (ii) and (iii), i.e., whether the transformation $G \mapsto G'$ is associated with a robust model response that allows resolutely crossing the decision boundary. Selecting any on-manifold transformation that flips the classifier (e.g., a minimal one) risks producing an adversarial example (cf.\ \cite{szegedy2016rethinking,DBLP:conf/cvpr/Stutz0S19}), as the counterfactual search may leverage spurious local variations in the classification function $f$ that are not representative of how $f$ varies more globally. Spurious local variations are commonplace in GNN classifiers \cite{Molspur} and also occur along the data manifold $\mathcal{M}$ \cite{DBLP:conf/cvpr/Stutz0S19}. Most existing counterfactual methods lack a mechanism to address potential spurious variations on the manifold.

\textbf{Non-Adversarial Rate (NA)}
To operationalize \textit{fidelity}, we evaluate whether counterfactuals occupy robust semantic basins rather than fragile adversarial pockets. Analogous to \cite{bender2025desiderata}, we validate the generated graphs against a retrained surrogate model $f_{sur}$ (distilled independently from $f$). This measures the transferability of the counterfactual flip, confirming it relies on robust structural features rather than weight-specific noise:
\begin{equation}
\text{Fidelity: NA} = \frac{N_{\text{flipped\_true}}}{N_{\text{flipped}}}
\end{equation}
where $N_{\text{flipped\_true}}$ represents the count of samples that also successfully flip the prediction of $f_{sur}$. Higher NA values indicate greater semantic stability and fidelity to the underlying decision process.

Further in the molecular domain, we will report \textbf{SMILES} ($\uparrow$), the fraction of counterfactuals that RDKit \cite{rdkit} can parse and sanitize under valence and bond-compatibility rules.  SMILES is defined only on the two molecular datasets, Mutagenicity and Benzene.

\subsection{Understandability}
\label{section:understandability}

Understandability refers to whether an explanation is understandable to its intended recipient, who is usually a human. Explanations should be presented at an appropriate level of abstraction and be concise enough to be quickly assimilated.
In the context of graph counterfactual explanations, an intrinsic level of understandability is guaranteed by the fact that generated counterfactuals live in the same input domain as factuals and reside on the data manifold $\mathcal{M}$ (i.e., inspectable discrete attributed graphs). Following \cite{bender2025desiderata}, we capture the concise nature required for understandability through structural \textit{sparseness} across the transformation $G \mapsto G'$.

To evaluate understandability fairly in the graph domain across different baselines is complicated by an asymmetry in their action spaces. Many baselines operate on a binary adjacency matrix and can only add or delete edges, whereas GDCE-I additionally edits atom and bond types. Reducing both to a single distance is therefore misleading. Further predicting the existence of a bond is chemically ambiguous (a single and a double bond between the same atoms are indistinguishable in a binary adjacency matrix), so any proximity computed on that representation abstracts away precisely the information that determines molecular identity. We therefore decompose structural minimality into three complementary sparsity metrics that jointly cover the attributed perturbation space while preserving comparability with edge-only baselines:

\textbf{Edge Sparsity.} Topological minimality measured as the fraction of added or removed edges relative to the original graph's edge set:
\begin{equation}
\text{Edge Sparsity}(G, G') = max(1 - \frac{|E \triangle E'|}{|E|}, 0)
\end{equation}
where $\triangle$ denotes the symmetric difference of the edge sets. The term $\frac{|E \triangle E'|}{|E|}$ is sometimes reported under the name `Modification Ratio' in related work papers, but we want to emphasize the connection between sparsity and understandability also elaborated in other interpretability research like~\cite{huben2024sparse}.

\textbf{Node Sparsity.} The fraction of node-type substitutions across all nodes, calculated relative to the node feature matrix:
\begin{equation}
\text{Node Sparsity}(G, G') = 1 - \frac{\sum_{i=1}^{|X|} \mathbb{1}[x_i \neq x_i']}{|X|}
\end{equation}

\textbf{Edge Type Sparsity.} Semantic changes on structurally preserved edges, measuring bond-type reassignments on retained edges:
\begin{equation}
\text{Edge Type Sparsity}(G, G') = 1 - \frac{|\{e \in E \cap E' \mid \text{attr}(e) \neq \text{attr}'(e)\}|}{|E \cap E'|}
\end{equation}

These three metrics isolate topological modifications, node-attribute substitutions, and edge-type shifts to provide a fine-grained quantification of concise, understandable edits. To illustrate the difference between fidelity (measured via NA) and understandability (measured via Sparsity), we provide practical examples in Table \ref{table:examples}.

\begin{table*}[t!]
\small \smallskip
\makebox[\textwidth][c]{
\begin{tabular}{cc}
\toprule
\parbox{4.5cm}{Faithful (High NA) but not Understandable (Low Sparsity)} & \parbox{11cm}{
\textit{A counterfactual that robustly flips both the target model and the surrogate model (high NA), but achieves this via an extensive graph-wide rewrite involving widespread topological rewiring, node substitutions, and edge-type modifications (poor Edge, Node, and Edge Type Sparsity) that overwhelm human inspection.}
}\\\midrule
\parbox{4.5cm}{Understandable (High Sparsity) but not Faithful (Low NA)} &
\parbox{11cm}{
\textit{A counterfactual with minimal structural edits (e.g., changing a single atom type or edge, yielding high Sparsity) that successfully flips the target classifier $f$, but fails to flip the surrogate model $f_{sur}$ (low NA) because it exploits a fragile, weight-specific adversarial vulnerability rather than a robust decision boundary.}
}\\\midrule
\parbox{4.5cm}{Faithful (High NA) but Insufficient (Low NAFR)} & \parbox{11cm}{
\textit{A counterfactual method that produces stable, non-adversarial edits when it succeeds (high NA), but fails to generate valid counterfactuals for a large fraction of the dataset, resulting in a low absolute Non-Adversarial Flip Rate (NAFR) and failing to globally map the classifier's behavior.}
}\\\bottomrule
\end{tabular}
}
\caption{Practical examples of graph counterfactuals that meet only some of the operationalized desiderata (NA, Sparsity, NAFR), adapting the conceptual framework of \cite{bender2025desiderata}.}
\label{table:examples}
\end{table*}
\subsection{Sufficiency}
\label{sec:sufficiency}

Following \cite{bender2025desiderata}, sufficiency requires charting the decision boundary globally across numerous distinct instances rather than exploiting localized or isolated successes. We operationalize sufficiency via the \textbf{Non-Adversarial Flip Rate (NAFR)}.

\textbf{Non-Adversarial Flip Rate (NAFR)}
Bridging both sufficiency and fidelity \cite{bender2025desiderata}, NAFR tracks the absolute proportion of generated samples across the dataset that simultaneously alter the target model's prediction and induce a genuine, transferable semantic shift confirmed by the surrogate classifier $f_{sur}$:
\begin{equation}
\text{Sufficiency:NAFR} = \frac{1}{N} \sum_{i=1}^{N} \mathbb{1}\left( f(G_i') = y_t \land f_{sur}(G_i') = y_t \right)
\end{equation}
By combining coverage and transferability into a single metric, NAFR isolates functional, structurally coherent graph counterfactuals from those that either fail to flip the target model or merely exploit imperceptible adversarial vulnerabilities.

\section{Method}

\paragraph{Discrete Denoising Diffusion on Graphs}
Let a graph be defined as $G = (X, E)$, where
$X \in \{0,1\}^{n \times a}$ represents the categorical node features
and $E \in \{0,1\}^{n \times n \times b}$ denotes the categorical edge features for $n$ nodes.

The fundamental objective of the forward diffusion process is to progressively transform an input graph $G \sim p_{data}$ into a sequence of increasingly noisy states, ultimately reaching a pure noise distribution $G^T$ at the final timestep $T$. While standard continuous diffusion models corrupt data by injecting Gaussian noise, applying this paradigm directly to adjacency matrices destroys the inherent sparsity of graphs, resulting in dense, fractional structures. To address this, discrete denoising diffusion models, such as DiGress \cite{DiGress}, define Markov transitions directly over categorical node and edge types via transition matrices $[Q^t_X]_{ij} = q(x^t=j|x^{t-1} = i)$ and $[Q^t_E]_{ij} = q(e^t=j|e^{t-1} = i)$. Similar to the continuous DDPM \cite{DDPM} forward process, the forward diffusion in discrete spaces is then defined as a Markovian process that iteratively adds noise at each timestep
\begin{equation}
q(G^t|G^{t-1}) = (X^{t-1}Q^t_X,E^{t-1}Q^t_E).
\label{eq:DiGress_forward}
\end{equation}

Due to the Markovian property, it is not necessary to apply noise recursively. By defining the cumulative transition matrix $\bar{Q}^t = Q^1Q^2 \dots Q^t$, we can directly sample the noisy graph at any arbitrary timestep $t$ from the initial clean graph $G$:
\begin{equation}
    q(G^t|G) = (X\bar{Q}^t_X,E\bar{Q}^t_E).
\label{eq:DiGress_forward_02}
\end{equation}
The goal of the generative model is to reverse this forward process. To do so, DiGress formulates the reverse process as a product of independent transitions over all nodes and edges

\begin{equation}
    p_\theta(G^{t-1}|G^t) = \prod_{1\leq i\leq n}p_\theta(x^{t-1}_i|G^t) \prod_{1\leq i,j\leq n}p_\theta(e^{t-1}_{ij}|G^t),
\label{eq:DiGress_reverse}
\end{equation}

where the individual marginals are computed by marginalizing over the clean state predictions $\hat{p}^X_i(x)$ produced by the denoising neural network

\begin{equation}
    p_\theta(x^{t-1}_i \mid G^t) = \sum_x \, q(x^{t-1}_i \mid x^t_i, x^0_i = x) \cdot \hat{p}^X_i(x)
\label{eq:DiGress_reverse_02}
\end{equation}

and $p_\theta(e^{t-1}_{ij} \mid G^t)$ is computed accordingly.
\paragraph{Classifier-Free guidance}
In conditional graph generation, traditional classifier-based guidance \cite{Classefier-Based} relies on an auxiliary property regressor evaluated on noisy intermediate graphs $G^t$ \cite{DiGress}. However, assigning meaningful continuous properties to such invalid, highly corrupted states is fundamentally ill-posed and yields unreliable guidance gradients. Classifier-Free Guidance (CFG) \cite{FreeGress, Classefier-Free} circumvents this issue by jointly training a single generative model for both conditional and unconditional generation. During training, the conditioning variable $y$ (e.g., a target class or continuous property) is randomly replaced by a null token $\emptyset$ with a predefined probability $p_{\textit{uncond}}$. During inference, the model extrapolates toward the conditional objective governed by a guidance scale $s$. 
Formally, the conditioned reverse process $p_\theta(G^{t-1}|G^t, y)$ is factorized over all nodes and edges under the assumption of conditional independence
 \begin{equation}
    p_\theta(G^{t-1}|G^t,y) = \prod_{1\leq i\leq n}p_\theta(x^{t-1}_i|G^t,y) \prod_{1\leq i,j\leq n}p_\theta(e^{t-1}_{ij}|G^t,y).
\label{eq:FreeGress_reverse}
\end{equation}
To compute the individual marginals, the model predicts the clean state from the noisy intermediate graph. For a node $i$, this marginalization is given by
\begin{equation}
    p_\theta(x^{t-1}_i \mid G_t,y) = \sum_x \, q(x^{t-1}_i \mid x^t_i, x^0_i = x) \cdot f_\theta(x^0_i=x|G^t,y)
\label{eq:FreeGress_reverse_02}
\end{equation}
where the clean state prediction $\hat{p}_\theta$ incorporates the classifier-free guidance mechanism. By linearly combining the conditional and unconditional predictions with the guidance scale $s$, the guided prediction is formulated as
\begin{equation}
    f_\theta(x^0_i=x|G^t,y) = p_\theta(x^0_i|G^t) + s\big(p_\theta(x^0_i|G^t, y) - p_\theta(x^0_i|G^t)\big)
\end{equation}

and the edge marginals $p_\theta(e^{t-1}_{ij} \mid G^t, y)$ are computed accordingly.

\paragraph{Categorical Sampling via the Gumbel-Max Trick}
\label{subsec:gumbel}
Both the forward and the reverse process require drawing samples from categorical distributions over node and edge types. Our inversion scheme relies on making this sampling step reparameterizable, so that its stochasticity can be recorded and later replayed. This is achieved with the Gumbel-Max trick. Let $p \in \Delta^{K-1}$ be a categorical distribution over $K$ classes, and let $g = (g_1, \dots, g_K)$ be a vector of independent standard Gumbel variables, $g_k \sim \mathrm{Gumbel}(0,1)$, which can be sampled as $g_k = -\log(-\log u_k)$ with $u_k \sim \mathcal{U}(0,1)$. Then
\begin{equation}
    \operatorname*{arg\,max}_{k}\big(\log p_k + g_k\big) \sim \mathrm{Categorical}(p),
\label{eq:gumbel_max}
\end{equation}
that is, perturbing the log-probabilities with Gumbel noise and taking the argmax yields an exact sample from $p$. Crucially, this decomposes a categorical draw into a deterministic component, the log-probabilities $\log p$ that depend on the conditioning, and a stochastic component, the noise $g$ that is independent of the content. Holding $g$ fixed while altering $\log p$ therefore changes the resulting sample only at positions where the shift in log-probabilities is large enough to move the argmax to a different class. This is the mechanism GDCE-I exploits to invert the discrete diffusion process in Section~\ref{sec:method}.

\subsection{GDCE-I}
\label{sec:method}
Building upon the theoretical foundations of discrete diffusion and classifier-free guidance, we present Graph Diffusion Counterfactual Explanations via Edit-Friendly Inversion (GDCE-I).
As a fundamental prerequisite to our framework, we first train a conditional discrete diffusion model on the target dataset. The conditioning signal utilized during this training phase can be derived directly from the predictions of the pre-trained black-box model, a process we term the distillation of the classifier into the discrete diffusion model. Notably, this conditional generative framework is highly versatile. The conditioning variables are not restricted to categorical classifier labels, but can seamlessly incorporate ground-truth variables, continuous physical properties (e.g., molecular logP), or even multi-dimensional property vectors. Once this conditional generative backbone is fully established, we can employ our inversion procedure to generate precise counterfactuals.

In continuous Denoising Diffusion Probabilistic Models (DDPMs), recent work has shown that the sequence of noise vectors drawn during the reverse process encodes the structural backbone of the generated sample, so that fixing this noise while altering the conditioning enables semantically meaningful edits that tightly preserve the original structure \cite{Edit_Friendly}. We translate this ``edit-friendly'' paradigm to the discrete graph domain. The central obstacle is that discrete diffusion samples from categorical transitions rather than adding Gaussian noise. We therefore realize an analogous noise space through the Gumbel-Max trick of Section~\ref{subsec:gumbel}, which separates each categorical draw into deterministic log-probabilities and content-independent Gumbel noise (Eq.~\ref{eq:gumbel_max}).

A naive realization would sample fresh Gumbel noise along the forward chain and hope the reverse process retraces it. This does not hold in general because the noise would be recorded against the forward transitions $Q^t$, yet replayed against the learned reverse posterior $p_\theta(\cdot \mid G^t, y)$, which is a different distribution. We instead construct an edit-friendly posterior noise space that guarantees exact reconstruction, using the truncated-Gumbel ($\mathrm{A}^\star$-sampling) construction \cite{AStarSampling}.

\paragraph{Reference trajectory} Given an input graph $G=(X,E)$ with its (distilled) condition $y$, we first draw a single reference trajectory $\hat{G}^0, \hat{G}^1, \dots, \hat{G}^T$ with $\hat{G}^0 = G$ by iterating the discrete forward process of Eq.~\ref{eq:DiGress_forward}, retaining only the visited states $\hat{G}^t = (\hat{X}^t, \hat{E}^t)$.

\paragraph{Posterior noise recording} We then traverse the trajectory in the reverse direction and, for every node and edge independently, record the Gumbel noise that forces the guided reverse posterior to reproduce the reference state. We describe the construction for a single categorical variable (one node; edges are analogous). At the reverse step from $t = s{+}1$ to $s$, let $\ell \in \mathbb{R}^{K}$ denote that node's guided reverse-posterior log-probabilities over the $K$ classes, $\ell = \log p_\theta(X^{s} \mid \hat{G}^{s+1}, y)$, and let $v^\star$ be the class realized at that node in the reference state $\hat{X}^{s}$. We seek Gumbel noise $g \in \mathbb{R}^{K}$ that selects this class under the Gumbel-Max rule of Eq.~\ref{eq:gumbel_max},
\begin{equation}
    \operatorname*{arg\,max}_{k}\big(\ell_k + g_k\big) = v^\star .
    \label{eq:posterior_target}
\end{equation}
Following the truncated-Gumbel ($\mathrm{A}^\star$-sampling) construction \cite{AStarSampling}, we build the perturbed log-probabilities $\phi_k = \ell_k + g_k$ directly. Let $Z = \log\sum_k e^{\ell_k}$ be the log-normalizer (here $Z = 0$, since $p_\theta$ is normalized) and let $M = Z + \gamma_0$ with $\gamma_0 \sim \mathrm{Gumbel}(0,1)$ be the maximal perturbed value. We assign $M$ to the target class and draw every other class as an independent $\mathrm{Gumbel}(\ell_k)$ value truncated to lie below $M$:
\begin{equation}
    \phi_k =
    \begin{cases}
        M, & k = v^\star, \\[2pt]
        -\log\!\big(e^{-(\ell_k + \gamma_k)} + e^{-M}\big), & k \neq v^\star,
    \end{cases}
    \qquad \gamma_k \sim \mathrm{Gumbel}(0,1).
    \label{eq:truncated_gumbel}
\end{equation}
By construction $\operatorname*{arg\,max}_k \phi_k = v^\star$. Since $\phi = \ell + g$, the reusable, condition-independent noise is recovered by subtracting the log-probabilities,
\begin{equation}
    g = \phi - \ell .
    \label{eq:noise_recover}
\end{equation}
Collecting $g$ over all nodes (resp.\ edges) yields the node noise $g_s^X$ (resp.\ the edge noise $g_s^E$, symmetrized across the adjacency). We store all recorded noise in a library $\mathcal{G} = \{(g_s^X, g_s^E)\}_{s=0}^{T-1}$ together with the reference states.

\paragraph{Guided generation.} To produce a counterfactual, we replay the reverse process with the same recorded noise but a new target condition $y'$. Stepping from $t$ to $t-1$, we compute the guided posterior $p_\theta(X^{t-1}\mid G^t, y')$ and sample deterministically with the stored noise,
\begin{equation}
    X^{t-1} = \mathrm{one\_hot}\Big(\operatorname*{arg\,max}\big(\log p_\theta(X^{t-1}\mid G^t, y') + g_{t-1}^X\big)\Big),
\end{equation}
and reconstruct $E^{t-1}$ symmetrically from $g_{t-1}^E$. By construction, replaying with the original condition ($y' = y$) recovers the input graph exactly. Under a changed condition $y'$, a position changes only where the guidance-induced shift in log-probabilities is large enough to move the argmax past the recorded margin. Edits are therefore concentrated precisely where the target property demands them, yielding sparse, in-distribution counterfactuals, and resolving the mismatch of the na\"ive scheme, whose noise is recorded and replayed against different distributions.  We show a comparison in \ref{subsec:ablation}.

\paragraph{Dynamic budget search.} Replaying the full trajectory from $T$ can introduce more changes than are needed to flip the classifier. We therefore introduce a budget $\tau \in \{1, \dots, T\}$: rather than replaying from $T$, we start the guided reverse pass from the reference state $\hat{G}^\tau$ and denoise down to $G^0$, reusing $\{(g_s^X, g_s^E)\}_{s < \tau}$. The framework searches over increasing $\tau$ and returns the counterfactual at the smallest budget for which $f(G^0) = y'$, which identifies the minimal structural deviation required to alter the prediction. Unlike DiGress, which draws plain categorical samples during generation \cite{DiGress}, GDCE-I's Gumbel-Max reparameterization is what renders the discrete trajectory invertible in the first place. The complete procedure, namely the reference trajectory, posterior recording, and budgeted guided replay, is summarized in Algorithm~\ref{alg:method} (Appendix~\ref{app:algorithm}).

\section{Experiments}
\label{sec:experiments}

To comprehensively evaluate the efficacy of our proposed framework, we design
our experiments to answer the following two primary research questions:
\begin{itemize}
    \item \textbf{Quantitative Benchmarking:} How does GDCE-I compare against
    state-of-the-art baseline explainers in terms of our defined evaluation framework on standard graph classification tasks?
    \item \textbf{Qualitative Analysis:} Can GDCE-I produce understandable edits?
\end{itemize}
\subsection{Experimental Setup}
\paragraph{Datasets}
For the discrete graph classification tasks, we evaluate our method on two widely adopted molecular benchmarks: Mutagenicity~\cite{mutagenicity} and Benzene \cite{BENZENE, BENZENXAI}. In these datasets, graphs represent chemical compounds. The binary label of Mutagenicity indicates a mutagenic effect, whereas Benzene is a structural attribution benchmark whose label indicates the presence of a benzene ring.

The focus on molecules is deliberate, and it concerns what can be measured rather than what the method supports. GDCE-I operates on categorical node and edge types and assumes nothing chemistry-specific. Molecular graphs, however, come with distributional constraints like valence rules. In-distribution behavior is therefore observable.

We additionally evaluate on two non-molecular benchmarks from the TUDataset collection~\cite{TUDataset}, both
without edge types. PROTEINS \cite{PROTEINS, PROTEINS01} represent proteins as graphs of
secondary-structure elements, with three categorical node labels (helix, sheet, turn)
and a binary label separating enzymes from non-enzymes. TWITTER \cite{TWITTER01, TWITTER} represents a tweet as a word co-occurrence graph, where nodes are words, and the binary label is the sentiment of
the tweet.

\paragraph{Baselines}
We benchmark GDCE-I against graph counterfactual explainers that produce instance-level counterfactuals for graph classification, spanning the heuristic/mask-based ($CF^2$~\cite{CF2}, C2Explainer~\cite{C2Explainer}, XPlore~\cite{XPlore}, UCExplainer~\cite{UCExplainer}) and generative (D4Explainer~\cite{D4Explainer}) families of Section~\ref{sec:related}. All baselines are re-evaluated under a single common protocol: identical dataset preprocessing, test split, sampling seeds, and metric definitions.
Crucially, every method is optimized against and evaluated by the same classifier. We train a binary-adjacency GCN, which every method can consume and against which all of them are measured, and additionally an edge-feature-aware GINE, reserved for the two methods that predict bond
types (GDCE-I and UCExplainer). Both are trained from scratch on the shared training split. Architectures and accuracies are given in appendix~\ref{app:datasets_classifiers}. PROTEINS and TWITTER carry no edge types and, therefore, have a GCN classifier only.

\paragraph{Counterfactuals evaluation}
\label{par:discrete_eval}
A counterfactual explanation must name a realizable modification of the input, so we
require the counterfactual to lie in the same space as the input: one-hot atom types,
one-hot bond types, and binary edges. Methods that optimize a continuous relaxation are
therefore discretized before any metric is computed. Node and bond vectors are projected back onto the simplex by $\arg\max$, soft edge masks are thresholded at $0.5$, and the Flip Rate is re-evaluated on the resulting graph. Two baselines are affected: UCExplainer returns node and bond vectors with several simultaneously active
entries, and D4Explainer evaluates its flip on sigmoid-weighted edges rather than on a
thresholded adjacency. For fair comparison, we report the post-discretisation value.
Further, $CF^2$, C2Explainer, XPlore and D4Explainer operate on a binary adjacency and predict no bond
types, so SMILES reconstruction requires a convention. $CF^2$ only deletes edges: every
surviving edge retains the bond order of the input, and no assignment is needed.
C2Explainer, XPlore and D4Explainer can additionally add edges, which carry no bond label. We assign these a single bond. This is the most frequent bond type in both datasets
($81.3\%$ of bonds in Mutagenicity, $56.3\%$ in Benzene) and, being the lowest-valence
option, the assignment is least likely to trigger a valence violation.

\subsection{Quantitative Evaluation on Graph Classification}

Table~\ref{tab:quantitative_results_main} reports every method against the same classifier per dataset, so
no difference between rows can be attributed to a differing classifier. For every method, we
evaluate five runs of $100$ molecules drawn from the test set under identical seeds. All
entries are means $\pm$ standard deviations over those runs. The only exception is the PROTEINS dataset, which is smaller, and we therefore evaluate on the whole test set.

GDCE-I attains the highest Flip Rate on all four datasets. This is
unchanged under NAFR, which additionally requires the flip to be confirmed by the control
judge, and GDCE-I likewise leads the SMILES
column on both datasets on which it is defined. It therefore makes the strongest case for altering the classifier's prediction while returning counterfactuals
that remain on the data manifold, and it does so while operating, together with UCExplainer, in the largest edit space of the compared methods.

UCExplainer attains the maximum edge
score of $1.000$ but just because it fools the classifier by adding entry in one-hot encoded vectors which the classifier never saw.
Among the methods that do edit topology, GDCE-I preserves the most of it on Mutagenicity, Benzene, and TWITTER. On PROTEINS, XPlore is ahead ($0.832$ vs $0.747$) but just with a lower NAFR $0.649$ against GDCE-I $0.747$. On atom types, GDCE-I leads outright on three of the four datasets.

\begin{table}[H]
\centering
\small
\setlength{\tabcolsep}{4pt}
\resizebox{\linewidth}{!}{
\begin{tabular}{l | cc | ccc | cc}
\toprule
\multicolumn{1}{c|}{} & \multicolumn{7}{c}{\cellcolor{CadetBlue!20}\textbf{desiderata}} \\
\cmidrule(l){2-8}
\multicolumn{1}{c|}{} & \multicolumn{2}{c|}{\cellcolor{CadetBlue!20}\textbf{sufficiency}} & \multicolumn{3}{c|}{\cellcolor{CadetBlue!20}\textbf{understandability}} & \multicolumn{2}{c}{\cellcolor{CadetBlue!20}\textbf{fidelity}} \\
\textbf{Method} & \cellcolor{CadetBlue!20}FR ($\uparrow$)& \multicolumn{1}{c|}{\cellcolor{CadetBlue!20}NAFR ($\uparrow$)} & \cellcolor{CadetBlue!20}EdgeSp. ($\uparrow$)& \cellcolor{CadetBlue!20}NodeSp. ($\uparrow$) & \multicolumn{1}{c|}{\cellcolor{CadetBlue!20}EdgeTypeSp. ($\uparrow$)} & \cellcolor{CadetBlue!20}NA ($\uparrow$)& \cellcolor{CadetBlue!20}SMILES ($\uparrow$) \\
\midrule
\multicolumn{8}{c}{\textbf{Mutagenicity}} \\
\midrule
$CF^2$~\cite{CF2} & $0.622_{\pm 0.029}$ & $0.510_{\pm 0.019}$ & $0.396_{\pm 0.020}$ & -- & -- & $0.820$ & $0.508_{\pm 0.016}$ \\
C2Explainer~\cite{C2Explainer} & $0.688_{\pm 0.016}$ & $0.558_{\pm 0.028}$ & $0.000^{\dagger}$ & -- & -- & $0.811$ & $0.002_{\pm 0.004}$ \\
XPlore~\cite{XPlore} & $0.922_{\pm 0.018}$ & $0.586_{\pm 0.069}$ & $0.613_{\pm 0.056}$ & -- & -- & $0.635$ & $0.050_{\pm 0.017}$ \\
UCExplainer (GCN)~\cite{UCExplainer} & $0.158_{\pm 0.044}$ & $0.148_{\pm 0.043}$ & $\mathbf{1.000}_{\pm 0.000}$ & $0.754_{\pm 0.036}$ & -- & $0.937$ & $0.088_{\pm 0.032}$ \\
UCExplainer (GINE) & $0.260_{\pm 0.044}$ & $0.186_{\pm 0.035}$ & $\mathbf{1.000}_{\pm 0.000}$ & $0.738_{\pm 0.050}$ & $0.910_{\pm 0.033}$ & $0.715$ & $0.068_{\pm 0.013}$ \\
D4Explainer~\cite{D4Explainer} & $0.138_{\pm 0.029}$ & $0.132_{\pm 0.033}$ & $0.597_{\pm 0.101}$ & -- & -- & $\mathbf{0.957}$ & $0.120_{\pm 0.040}$ \\
\textbf{GDCE-I (Ours, GCN)} & $\mathbf{0.970}_{\pm 0.016}$ & $\mathbf{0.886}_{\pm 0.029}$ & $0.657_{\pm 0.011}$ & $0.892_{\pm 0.009}$ & $0.993_{\pm 0.000}$ & $0.913$ & $\mathbf{0.744}_{\pm 0.017}$ \\
\textbf{GDCE-I (Ours, GINE)} & $0.770_{\pm 0.049}$ & $0.628_{\pm 0.068}$ & $0.680_{\pm 0.016}$ & $\mathbf{0.919}_{\pm 0.006}$ & $\mathbf{0.994}_{\pm 0.002}$ & $0.816$ & $0.534_{\pm 0.069}$ \\
\midrule
\multicolumn{8}{c}{\textbf{Benzene}} \\
\midrule
$CF^2$ & $0.566_{\pm 0.021}$ & $0.550_{\pm 0.007}$ & $0.391_{\pm 0.005}$ & -- & -- & $0.972$ & $0.000_{\pm 0.000}$ \\
C2Explainer & $0.666_{\pm 0.038}$ & $0.610_{\pm 0.032}$ & $0.910_{\pm 0.010}$ & -- & -- & $0.916$ & $0.092_{\pm 0.022}$ \\
XPlore & $0.670_{\pm 0.021}$ & $0.598_{\pm 0.004}$ & $0.780_{\pm 0.038}$ & -- & -- & $0.893$ & $0.206_{\pm 0.036}$ \\
UCExplainer (GCN) & $0.202_{\pm 0.023}$ & $0.184_{\pm 0.029}$ & $\mathbf{1.000}_{\pm 0.000}$ & $0.818_{\pm 0.027}$ & -- & $0.911$ & $0.092_{\pm 0.022}$ \\
UCExplainer (GINE) & $0.558_{\pm 0.029}$ & $0.508_{\pm 0.036}$ & $\mathbf{1.000}_{\pm 0.000}$ & $0.459_{\pm 0.020}$ & $0.369_{\pm 0.014}$ & $0.910$ & $0.000_{\pm 0.000}$ \\
D4Explainer & $0.290_{\pm 0.054}$ & $0.290_{\pm 0.054}$ & $0.597_{\pm 0.080}$ & -- & -- & $\mathbf{1.000}$ & $0.290_{\pm 0.054}$ \\
\textbf{GDCE-I (Ours, GCN)} & $0.888_{\pm 0.037}$ & $0.858_{\pm 0.046}$ & $0.915_{\pm 0.011}$ & $\mathbf{0.910}_{\pm 0.004}$ & $\mathbf{0.998}_{\pm 0.001}$ & $0.966$ & $\mathbf{0.360}_{\pm 0.044}$ \\
\textbf{GDCE-I (Ours, GINE)} & $\mathbf{0.948}_{\pm 0.020}$ & $\mathbf{0.938}_{\pm 0.026}$ & $0.947_{\pm 0.007}$ & $0.882_{\pm 0.005}$ & $0.989_{\pm 0.003}$ & $0.989$ & $0.286_{\pm 0.039}$ \\
\midrule
\multicolumn{8}{c}{\textbf{PROTEINS}} \\
\midrule
$CF^2$ & $0.759$ & $0.730$ & $0.523$ & -- & -- & $\mathbf{0.962}$ & -- \\
C2Explainer & $0.431$ & $0.374$ & $0.365$ & -- & -- & $0.868$ & -- \\
XPlore & $0.805$ & $0.649$ & $0.875$ & -- & -- & $0.807$ & -- \\
UCExplainer (GCN) & $0.241$ & $0.224$ & $\mathbf{1.000}$ & $0.773$ & -- & $0.929$ & -- \\
D4Explainer & $0.132$ & $0.121$ & $0.832$ & -- & -- & $0.917$ & -- \\
\textbf{GDCE-I (Ours, GCN)} & $\mathbf{0.851}$ & $\mathbf{0.747}$ & $0.762$ & $\mathbf{0.872}$ & -- & $0.878$ & -- \\
\midrule
\multicolumn{8}{c}{\textbf{TWITTER}} \\
\midrule
$CF^2$ & $0.194_{\pm 0.019}$ & $0.130_{\pm 0.027}$ & $0.213_{\pm 0.037}$ & -- & -- & $0.670$ & -- \\
C2Explainer & $0.054_{\pm 0.021}$ & $0.036_{\pm 0.018}$ & $0.313_{\pm 0.133}$ & -- & -- & $0.667$ & -- \\
XPlore & $0.142_{\pm 0.029}$ & $0.082_{\pm 0.016}$ & $0.621_{\pm 0.049}$ & -- & -- & $0.582$ & -- \\
UCExplainer (GCN) & $0.532_{\pm 0.069}$ & $0.474_{\pm 0.061}$ & $\mathbf{1.000}_{\pm 0.000}$ & $\mathbf{0.522}_{\pm 0.066}$ & -- & $0.891$ & -- \\
D4Explainer & $0.000_{\pm 0.000}$ & $0.000_{\pm 0.000}$ & -- & -- & -- & -- & -- \\
\textbf{GDCE-I (Ours, GCN)} & $\mathbf{0.996}_{\pm 0.005}$ & $\mathbf{0.954}_{\pm 0.015}$ & $0.957_{\pm 0.009}$ & $0.496_{\pm 0.032}$ & -- & $\mathbf{0.958}$ & -- \\
\bottomrule
\end{tabular}}
\caption{All metrics are oriented so that higher is better ($\uparrow$);
subscripts give the standard deviation over five runs of $100$ graphs. The three sparsity columns are $1$ minus the corresponding modification
rate, so $1$ means that part of the graph was left untouched. SMILES is defined on the molecular datasets
only. $^{\dagger}$clamped at $0$; the raw modification rate is $1.390$.
PROTEINS is a single run over the complete test split and admits no spread.}
\label{tab:quantitative_results_main}
\end{table}

GDCE-I also allocates its edit budget to the property that defines each dataset. On
Mutagenicity, it spends the budget on topology, preserving $0.892$ of the atom types but
only $0.657$ of the edges, consistent with rearranging the substructures that carry
mutagenicity. On Benzene, where the label is the presence of a benzene ring, it instead
exploits an edit that no edge-only baseline can express and substitutes a ring carbon by
another atom type. The ordering reverses, with $0.947$ of the edges but only $0.882$ of the atom types left intact.

GDCE-I does not lead the NA column on the molecular datasets but stays competitive. One should note that NA is conditioned on the flips a method produces. D4Explainer's $1.000$ on Benzene is computed over the $29.0\%$ of graphs it flips at all, whereas GDCE-I's $0.966$ is computed over $88.8\%$. Read together
with NAFR, which shares its denominator with FR, the two columns separate how often a method flips from how often its flips survive a change of judge.

\paragraph{The target classifier is not a neutral choice}
Because the shared protocol is what makes Table~\ref{tab:quantitative_results_main} comparable, it is worth stating how much it changes. Re-running each baseline against \textit{its own} paper
oracle on the same splits, seeds, and metric definitions, exchanging nothing but the
classifier, moves the Flip Rate by up to $0.30$ on Mutagenicity: C2Explainer falls from $0.984$
to $0.688$, D4Explainer from $0.296$ to $0.154$ and $CF^2$ from $0.820$ to $0.622$. The
effect is not uniform in sign, as UCExplainer gains under the shared judge.
The largest losses fall on precisely those methods that
report a near-perfect Flip Rate against their own oracle, whereas GDCE-I moves by at most
$0.06$ in either direction ($0.908 \to 0.970$ on Mutagenicity).

Sparsity behaves differently. For most methods, structural cost is close to classifier-invariant: $CF^2$
($0.600 \to 0.604$), D4Explainer ($0.429 \to 0.384$) and GDCE-I ($0.344 \to 0.343$) report essentially the same Edge Sparsity on Mutagenicity under either classifier. C2Explainer is the exception, rising from $0.084$ to $1.390$, reproduced as $15.3\times$ in a paired control that exchanges nothing but the classifier.

\subsection{Ablation Studies}
\label{subsec:ablation}

\paragraph{The role of the inversion}
We compare three different sampling schemes under the same training run, same checkpoint, same shared GCN classifier, guidance scale $s=3$, same dynamic budget search, and $5\times100$ Mutagenicity samples.
Firstly \textbf{no inversion}, which perturbs the source to $G_\tau$ along the forward trajectory and then samples the reverse process freshly. Secondly \textbf{naive inversion}, which records
and re-injects a fresh Gumbel sample rather than the posterior-consistent one, and \textbf{posterior inversion} (GDCE-I), which re-injects the exact edit-friendly noise \ref{sec:method}.
Table~\ref{tab:ablation_inversion} reports the outcome under the same metrics.

\begin{table}[H]
\centering
\small
\setlength{\tabcolsep}{4pt}
\resizebox{\linewidth}{!}{
\begin{tabular}{l | cc | ccc | cc | c}
\toprule
\multicolumn{1}{c|}{} & \multicolumn{7}{c|}{\cellcolor{CadetBlue!20}\textbf{desiderata}} & \\
\cmidrule(lr){2-8}
\multicolumn{1}{c|}{} & \multicolumn{2}{c|}{\cellcolor{CadetBlue!20}\textbf{sufficiency}} & \multicolumn{3}{c|}{\cellcolor{CadetBlue!20}\textbf{understandability}} & \multicolumn{2}{c|}{\cellcolor{CadetBlue!20}\textbf{fidelity}} & \\
\textbf{Variant} & \cellcolor{CadetBlue!20}FR ($\uparrow$) & \multicolumn{1}{c|}{\cellcolor{CadetBlue!20}NAFR ($\uparrow$)} & \cellcolor{CadetBlue!20}EdgeSp. ($\uparrow$) & \cellcolor{CadetBlue!20}NodeSp. ($\uparrow$) & \multicolumn{1}{c|}{\cellcolor{CadetBlue!20}EdgeTypeSp. ($\uparrow$)} & \cellcolor{CadetBlue!20}NA ($\uparrow$) & \multicolumn{1}{c|}{\cellcolor{CadetBlue!20}SMILES ($\uparrow$)} & $\tilde{\tau}$ \\
\midrule
No inversion & $\mathbf{1.000}_{\pm 0.000}$ & $0.876_{\pm 0.027}$ & $\mathbf{0.675}_{\pm 0.028}$ & $\mathbf{0.923}_{\pm 0.008}$ & $\mathbf{0.997}_{\pm 0.001}$ & $0.876_{\pm 0.027}$ & $0.804_{\pm 0.025}$ & $11$ \\
Naive inversion & $0.996_{\pm 0.005}$ & $0.858_{\pm 0.019}$ & $0.592_{\pm 0.033}$ & $0.907_{\pm 0.007}$ & $0.995_{\pm 0.002}$ & $0.861_{\pm 0.018}$ & $\mathbf{0.832}_{\pm 0.036}$ & $36$ \\
\textbf{Posterior (GDCE-I)} & $0.970_{\pm 0.016}$ & $\mathbf{0.886}_{\pm 0.029}$ & $0.657_{\pm 0.011}$ & $0.892_{\pm 0.009}$ & $0.993_{\pm 0.000}$ & $\mathbf{0.913}_{\pm 0.029}$ & $0.816_{\pm 0.021}$ & $51$ \\
\bottomrule
\end{tabular}}
\caption{Inversion ablation on Mutagenicity against the shared GCN classifier
($5\times100$, $s=3$), reported under the metrics of
Table~\ref{tab:quantitative_results_main}; $\tilde{\tau}$ is the median flip budget. All
three variants come from the same training run and checkpoint and differ only in the noise
treatment, and the posterior row is the GDCE-I (GCN) row of
Table~\ref{tab:quantitative_results_main}. The column-wise best is in \textbf{bold}.}
\label{tab:ablation_inversion}
\end{table}

All three are strong counterfactual generators, and no desideratum separates them. They flip
almost every instance ($1.000$, $0.996$, $0.970$), and once the flip has to survive the
control judge, they stay within one standard deviation of one another ($0.876$, $0.858$,
$0.886$). The joint criterion of being both a flip and a molecule falls within $0.03$ ($0.804$, $0.832$,
$0.816$). On the understandability axes, the spread reaches at most two standard deviations:
no inversion preserves the most topology ($0.675$ against $0.657$ and the most atom types ($0.923$ against $0.892$ for posterior inversion.

What the columns separate is how the flip is obtained, and the flip budget makes it
visible. Without noise recording, the reverse process is freshly resampled, diverges almost immediately from
the source, and finds a flip at a median budget of $\tilde{\tau}=11$.
Posterior inversion reconstructs the reference exactly at small $\tau$ and therefore cannot
flip until the budget grows to $\tilde{\tau}=51$. Naive inversion, which re-injects noise that the reverse posterior was never conditioned on, sits between the two at $36$. The choice is thus a trade-off rather than a ranking. Posterior inversion is the only scheme
that reconstructs, whereas no inversion reaches a flip at roughly a fifth of the budget.

\begin{figure}[H]
    \centering
    \includegraphics[width=\textwidth]{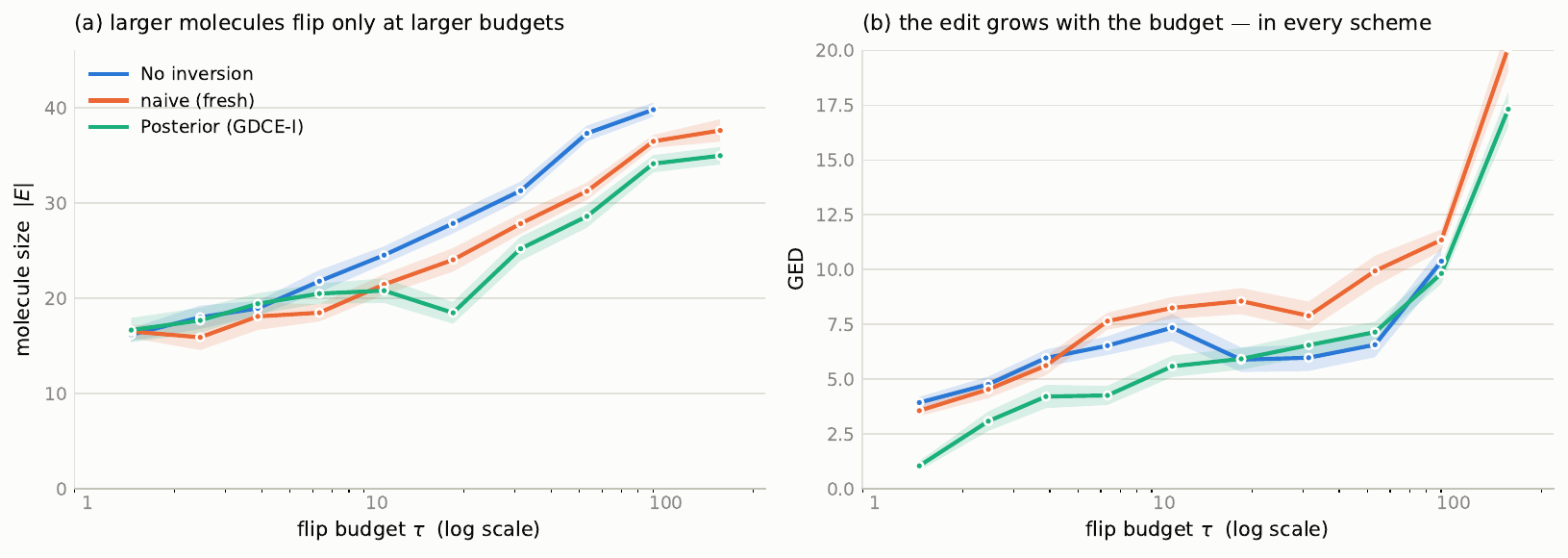}
    \caption{Flip budget $\tau$ on the horizontal axis of both panels (Mutagenicity, shared
    GCN classifier, $s=3$, successful counterfactuals; binned means $\pm 1$ SEM, log scale).
    \textit{(a)} Moleculesize against $\tau$.
    \textit{(b)} Absolute edit size against $\tau$.}
    \label{fig:tau_sparsity}
\end{figure}

\paragraph{Budget, molecule size, and edit size}
The budget search assumes that a smaller $\tau$ yields a smaller edit. Testing that assumption, we observe (Figure ~\ref{fig:tau_sparsity}) that this holds for all variants and also that the initial graph size sets the tone of how much budget is needed to find the counterfactual.

Posterior inversion lies below both alternatives at every budget. This is what exact reconstruction buys at the same budget on the same molecule. The guided replay departs from the source itself rather than from somewhere near it.

\paragraph{Guidance strength.}
The guidance scale $s$ governs how forcefully generation is steered toward the target
class, and sweeping it over $s \in \{1,3,5\}$ separates two effects that move in opposite
directions. Raising $s$ improves every axis that measures success and economy at once. The
Flip Rate ($0.740 \to 0.970 \to 0.998$), all three understandability axes ($0.606 \to 0.657
\to 0.726$ on edges, $0.882 \to 0.892 \to 0.905$ on atom types, $0.991 \to 0.993 \to 0.995$
on bond types) and the flip budget ($\tilde{\tau}$ $86 \to 51 \to 25$). Stronger guidance reaches the target class earlier along the replay and, therefore, departs less far from the source.

Fidelity moves the other way. The rate at which a flip is transmitted to the control judge decreases with $s$ ($0.944 \to 0.913 \to 0.879$), and so does the share of counterfactuals that RDKit can sanitize, conditioned on a flip ($0.852 \to 0.841 \to 0.776$). Both point at
the same mechanism: forcing the classifier signal harder buys a high flip rate and sparsity, but pushes the sample off the data manifold. 

A counterfactual is only useful if it fulfills all the requirements at once. The two columns that combine these requirements the best are in the middle. The joint flip-and-molecule rate at $s=3$ ($0.816$
against $0.630$ and $0.774$) and the non-adversarial flip rate likewise ($0.886$ against
$0.699$ and $0.879$). We adopt $s=3$ for the main results.
\begin{table}[H]
\centering
\small
\setlength{\tabcolsep}{4pt}
\resizebox{\linewidth}{!}{
\begin{tabular}{c | cc | ccc | cc | c}
\toprule
\multicolumn{1}{c|}{} & \multicolumn{7}{c|}{\cellcolor{CadetBlue!20}\textbf{desiderata}} & \\
\cmidrule(lr){2-8}
\multicolumn{1}{c|}{} & \multicolumn{2}{c|}{\cellcolor{CadetBlue!20}\textbf{sufficiency}} & \multicolumn{3}{c|}{\cellcolor{CadetBlue!20}\textbf{understandability}} & \multicolumn{2}{c|}{\cellcolor{CadetBlue!20}\textbf{fidelity}} & \\
$s$ & \cellcolor{CadetBlue!20}FR ($\uparrow$) & \multicolumn{1}{c|}{\cellcolor{CadetBlue!20}NAFR ($\uparrow$)} & \cellcolor{CadetBlue!20}EdgeSp. ($\uparrow$) & \cellcolor{CadetBlue!20}NodeSp. ($\uparrow$) & \multicolumn{1}{c|}{\cellcolor{CadetBlue!20}EdgeTypeSp. ($\uparrow$)} & \cellcolor{CadetBlue!20}NA ($\uparrow$) & \multicolumn{1}{c|}{\cellcolor{CadetBlue!20}SMILES ($\uparrow$)} & $\tilde{\tau}$ \\
\midrule
$1$ & $0.740_{\pm 0.037}$ & $0.699_{\pm 0.029}$  & $0.606_{\pm 0.027}$ & $0.882_{\pm 0.009}$ & $0.991_{\pm 0.003}$ & $\mathbf{0.944}_{\pm 0.054}$  & $0.630_{\pm 0.042}$ & $86$ \\
$\mathbf{3}$ & $0.970_{\pm 0.016}$ & $\mathbf{0.886}_{\pm 0.029}$ & $0.657_{\pm 0.011}$ & $0.892_{\pm 0.009}$ & $0.993_{\pm 0.000}$ & $0.913_{\pm 0.029}$ & $\mathbf{0.816}_{\pm 0.021}$ & $51$ \\
$5$ & $\mathbf{0.998}_{\pm 0.004}$ & $\mathbf{0.894}_{\pm 0.023}$ & $\mathbf{0.726}_{\pm 0.028}$ & $\mathbf{0.905}_{\pm 0.005}$ & $\mathbf{0.995}_{\pm 0.003}$ & $0.896_{\pm 0.020}$  & $0.774_{\pm 0.025}$ & $24$ \\
\bottomrule
\end{tabular}}
\caption{Guidance-scale sweep for GDCE-I (posterior inversion) on Mutagenicity against the
shared GCN classifier ($5\times100$), reported under the metrics of
Table~\ref{tab:quantitative_results_main}; $\tilde{\tau}$ is the median flip budget. Column-wise best in \textbf{bold}.}
\label{tab:ablation_sweep}
\end{table}

\subsection{Qualitative Analysis on Mutagenicity}

To evaluate whether the generated counterfactuals provide meaningful insights, we conducted a qualitative analysis on the Mutagenicity dataset. This reference comprises aromatic and heteroaromatic chemical compounds classified according to their mutagenic effect on the Gram-negative bacterium Salmonella typimurium~\cite{mutagenicity}. Toxicophore derivation studies have identified multiple functional subgroups that are highly discriminatory for the classification of mutagenicity \cite{mutagenicity}. Among these, the aromatic nitro group, particularly the benzene-$\mathrm{NO_2}$ motif, emerges as the most frequently occurring and highly predictive structural characteristic. Kazius et al.~\cite{mutagenicity} report that $87\%$ of the molecules containing this substructure are mutagens, a finding mirrored by our shared GCN classifier, which predicts $94.6\%$ of the $669$ molecules that carry this motif as mutagenic across the combined train, validation, and test sets.

The qualitative evaluation of counterfactual methods on this motif was first established by the authors of CF2 \cite{CF2}, who constructed a filtered subset called ``Mutag0'' \cite{CF2}. Their analysis showed that purely deletion-based methods tend to simply remove a few critical edges to break the toxicophore, which fails to yield an explanation \cite{CF2} by ending up in a fragmented counterfactual.
However, despite identifying this flaw, CF2 itself suffers from similar limitations. As a mask-based attribution method, CF2 is designed primarily to identify and extract an existing factual subgraph rather than generate a counterfactual instance. Because its counterfactual reasoning strictly defines the counterfactual as the original graph minus the identified factual mask, CF2 is restricted to edge and node deletions. Consequently, while the method successfully highlights the relevant toxicophore, it will not necessarily generate a counterfactual, which is an instance of the data manifold.

To evaluate whether GDCE-I is capable of solving these issues, we design a targeted experiment. We filter the test set to isolate molecules that contain both the benzene-$\mathrm{NO_2}$ motif and are explicitly classified as mutagenic by the GNN classifier. Then we apply our framework to generate counterfactual explanations for this subset. A faithful counterfactual explainer should naturally recognize the causal significance of this motif and specifically alter the aromatic nitro group to flip the prediction from mutagenic to non-mutagenic.

\begin{figure}[H]
    \centering
    \includegraphics[width=\textwidth]{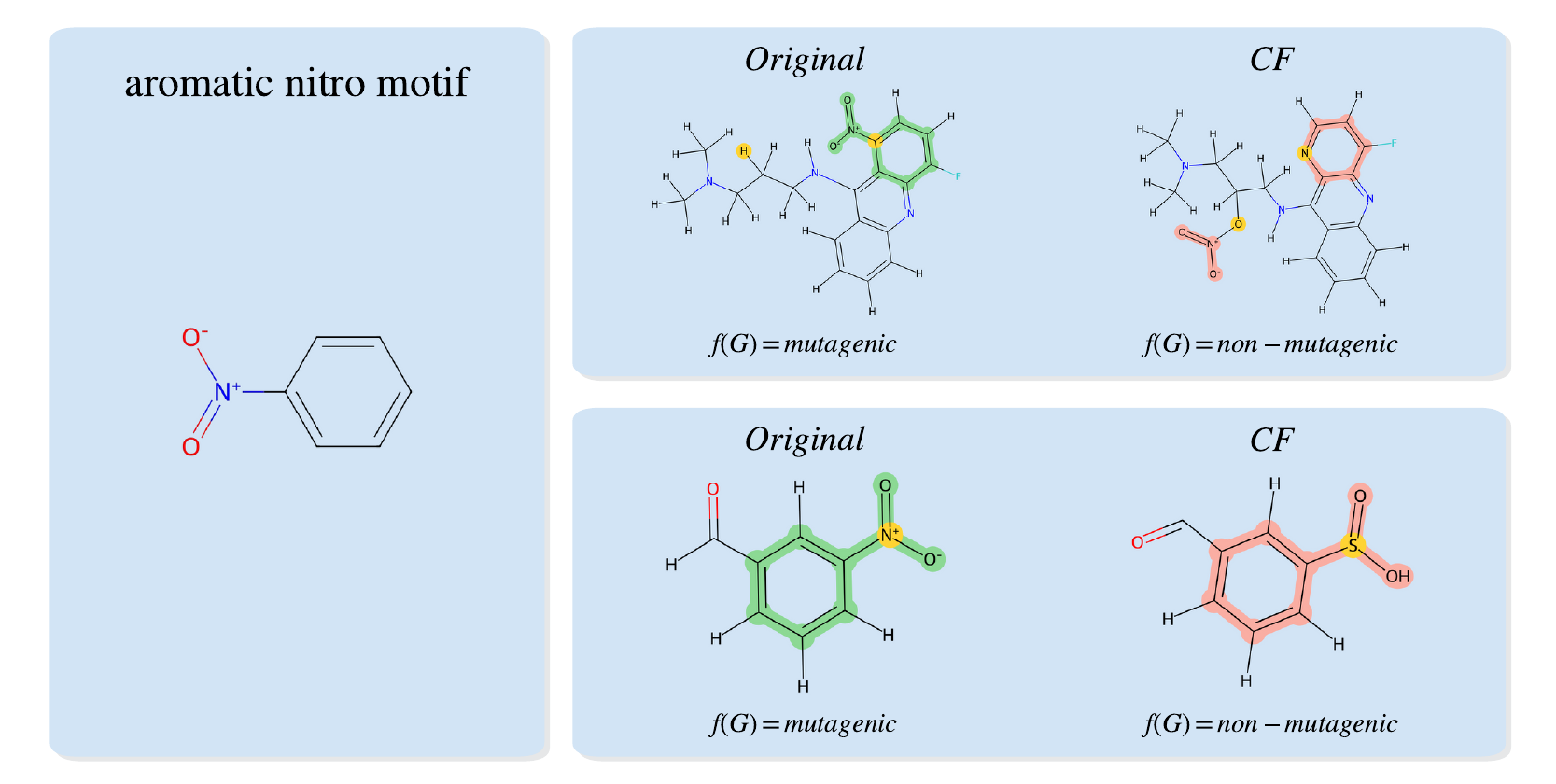}
    \caption{Qualitative counterfactuals on Benzene (GDCE-I (GCN), $s=3$).Highlighting: \textit{green} marks the original motif, \textit{red}
    the remainings, and \textit{yellow} the edits. \textit{(Left)} The aromatic nitro motif. \textit{(Top right)}: The original mutagenic compound carries an $\mathrm{NO_2}$ group connected to an aromatic ring.
    The generated counterfactual detoxifies the molecule by relocating the nitro group from the aromatic to a non-aromatic structural position.
    \textit{(Bottom right)}: GDCE-I performs a direct functional group substitution, replacing the mutagenic $\mathrm{NO_2}$ group with a, non-mutagenic $\mathrm{SO_2}$ group.}
    \label{fig:figure_mutagenicity}
\end{figure}

Our empirical evaluation strongly validates this expectation. Of the $116$ molecules in this targeted subset, GDCE-I alters the prediction of the classifier for $115$ ($99.1\%$). A detailed structural analysis further reveals that in $93.0\%$ of the generated
counterfactuals, the explainer explicitly modifies the benzene-$\mathrm{NO_2}$ motif. $72.2\%$ of the counterfactuals in this subset yield a valid SMILE. Figure~\ref{fig:figure_mutagenicity} presents two representative examples. In the first, the method detoxifies the molecule by detaching the $\mathrm{NO_2}$ group from the benzene ring and relocating it to a non-aromatic structural position. In the second, it performs a direct substitution of functional groups, replacing the mutagenic $\mathrm{NO_2}$ group with a benign $\mathrm{SO_2}$ group. Both strategies neutralize the primary toxicophore while strictly preserving the topological backbone of the original molecule, underscoring the capability of GDCE-I to deliver understandable counterfactuals.

\subsection{Qualitative Analysis on Benzene}

We conduct another test on Benzene. A graph is exactly positive when it contains a
benzene ring~\cite{BENZENE, BENZENXAI}. Thus, a faithful explanation would destroy the benzene ring when the target class is negative and construct one when it is positive.

GDCE-I flips $94.8\%$ of the sampled molecules (Table~\ref{tab:quantitative_results_main}). Among
the counterfactuals that produce a valid SMILES, $86.4\%$ satisfy the
ground-truth motif condition, i.e., \ the benzene ring is actually removed or created.
In the positive-to-negative direction, this holds for every counterfactual.
Crucially, $74.9\%$ of all flips are obtained without inserting or deleting a single
edge. GDCE-I breaks or completes the ring by substituting a ring atom, an edit
that leaves the surrounding topology untouched. 

\begin{figure}[H]
    \centering
    \includegraphics[width=\textwidth]{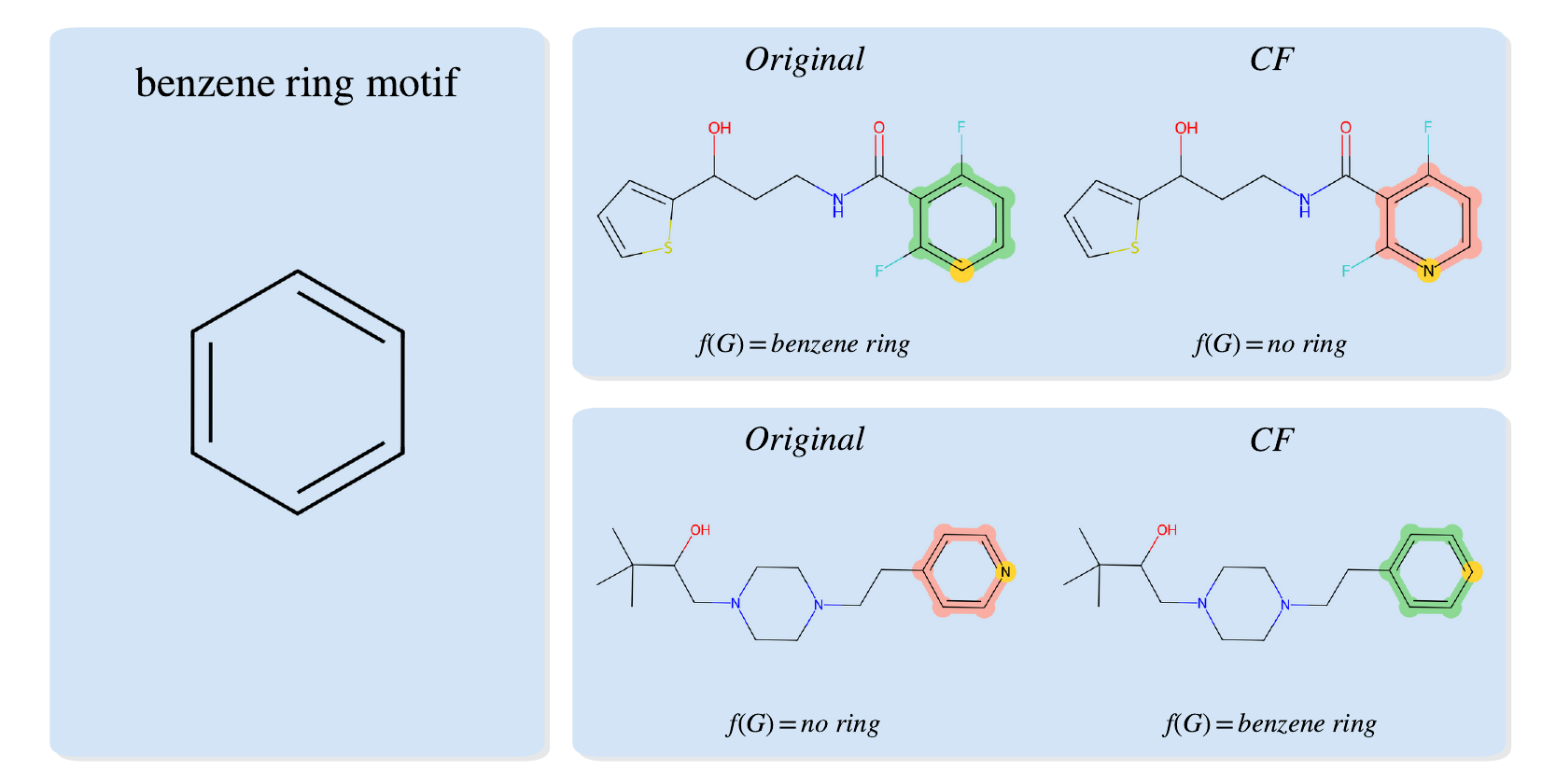}
    \caption{Qualitative counterfactuals on Benzene (GDCE-I (GINE), $s=3$).
    Highlighting: \textit{green} marks the original motif, \textit{red}
    the remainings, and \textit{yellow} the edits.
    \textit{(Left)} The benzene-ring motif.
    \textit{(Top right)}: A ring carbon of the original is replaced by nitrogen, flipping the prediction to no benzene ring. \textit{(Bottom right)}: The reverse direction, where a nitrogen is replaced by carbon so that a benzene ring is completed.}
    \label{fig:figure_benzene}
\end{figure}

Figure~\ref{fig:figure_benzene} shows one representative example per direction, both
 single-atom substitutions with Edge Sparsity and Edge Type Sparsity of exactly
$1$. In the Top right, carbon is replaced by nitrogen. Notably, the thiophene at the other end of the molecule, which is aromatic but not a benzene ring, is left untouched, so the edit targets the labeled motif rather than aromaticity in general. The second example runs the other way: a pyridine nitrogen is replaced by carbon, completing a benzene ring and flipping the prediction to the positive class. Together with the Mutagenicity analysis, this shows
that GDCE-I localizes the class-defining substructure, and edits it data-manifold aware.

\section{Limitations}
\label{sec:limitations}

Our results should not be read as evidence that generative counterfactual explainers dominate heuristic ones. What a distribution-aware generator buys is that its edits are drawn from a model which learned the data distribution. Molecular graphs constrain it sharply, but on benchmarks whose distribution is not that unconstrained, such as synthetic graphs built from planted motifs, we expect modeling the data manifold buys nothing while its cost remains, and we would not expect GDCE-I to be the appropriate tool.
Further, GDCE-I and other generative approaches require a conditional diffusion model trained per dataset, a cost the mask-based baselines do not incur at all. This rules the method out where an explanation is needed on a dataset for which no generative model exists, or it's too hard to be trained, a regime in which mask-based explainers remain structurally advantaged regardless of the quality of their edits.

The counterfactuals we generate are edits to a molecular graph. Atom types, bond types, and topology. Many of the questions that motivate counterfactual reasoning in chemistry, from binding to pharmacokinetics, are not decided at that level, but by the three-dimensional conformations a compound adopts. A topological edit that flips a classifier trained on graphs therefore states a hypothesis about the molecule, not about its behavior in a conformation space.
\section{Conclusion}

We presented GDCE-I, a framework for graph counterfactual explanation built on discrete
diffusion inversion. Its central component is a Gumbel-Max inversion that records the
sampling noise of a reference trajectory in closed form, so that replaying the reverse
process under the original condition reconstructs the input exactly, and replaying it under
a changed condition alters only those positions where guidance is strong enough to move the
argmax past the recorded margin. The budget parameter $\tau$ turns minimality into a search
over how much of the trajectory is regenerated rather than a penalty term fixed at training
time.

This addresses the two problems we identified. Because every edit is a step of a discrete
diffusion model of the data, the counterfactual is held on the data manifold rather than
merely close to the input. Secondly, because that model is defined over node types, bond types, and topology within a single generative process, the full edit space can be leveraged to find an explanation. The results follow from both. GDCE-I attains the highest
Flip Rate of all compared methods on every benchmark ($0.970$ on Mutagenicity, $0.948$ on
Benzene, $0.851$ on PROTEINS, $0.996$ on TWITTER) and the highest non-adversarial flip rate
throughout ($0.886$, $0.938$, $0.747$, $0.954$). On the two molecular benchmarks, where membership of the data manifold can actually be tested, it returns the largest share of counterfactuals that are at once non-adversarial and a valid molecule ($0.744$ and $0.360$).

Our second contribution is the evaluation itself. The metrics reported in this area are incomplete and inconsistent: the same names denote different amounts in papers, and different evaluation schemes are applied. We therefore derived desiderata for counterfactual explanation and defined suitable metrics for the graph domain. 

\section{Future Work}

One possible direction of future work might be to also introduce diversity in the counterfactual search as done for tabular data, e.g., in~\cite{mothilal2020explaining} or for visual classifiers in~\cite{bender2025desiderata}.
This might enable us to systematically find secondary Clever Hans features~\cite{CleverHans} and remove them, e.g., with CFKD~\cite{bender2023towards, bender2026mitigating} by adapting it plug-and-play from vision and tabular counterfactuals to graph counterfactuals.
Another direction to explore would be applying the Gumbel softmax trick for editing with discrete diffusion to other inherently discrete domains like natural language or protein sequences.

\subsubsection*{Acknowledgments}
We thank Klaus-Robert Müller for general discussion and guidance. We also thank Stefan Gugler for help and guidance for the qualitative experiments on Mutagenicity and Benzen. 
This work was partly funded by the German Ministry for Education and Research (under refs 01IS14013A-E, 01GQ1115, 01GQ0850, 01IS18056A, 01IS18025A, 13GW0744D, and BIFOLD25B) and by  DFG. Sidney Bender was partially funded by Bosch-Siemens Haushaltsger\"ate.

\bibliography{references}

\appendix
\section{Full Algorithm}
\label{app:algorithm}

\begin{algorithm}[H]
\small
\SetAlgoLined
\KwIn{Clean graph $G = (X, E, y)$, target $y'$, GNN classifier $f$, budget schedule $\mathcal{T}_{skip}$}
\KwOut{$G_{CE} = (X_{CE}, E_{CE}, y')$}

\tcp{Phase 1: reference trajectory (states only)}
$\hat{G}^0 \leftarrow G$\;
\For{$t = 1$ \KwTo $T$}{
    $\hat{X}^t \sim \hat{X}^{t-1}Q^t_X, \quad \hat{E}^t \sim \hat{E}^{t-1}Q^t_E$ \tcp*[r]{single-step forward, Eq.~\ref{eq:DiGress_forward}}
}

\vspace{0.15cm}
\tcp{Phase 2: posterior Gumbel recording (reverse, original condition $y$)}
$\mathcal{G} \leftarrow \emptyset$\;
\For{$s = T-1$ \KwTo $0$}{
    $p_\theta(x^{s}_i \mid \hat{G}^{s+1}, y) \leftarrow \sum_x q(x^{s}_i \mid \hat{x}^{s+1}_i, x^0_i = x)\, f_\theta(x^0_i=x \mid \hat{G}^{s+1}, y)$\;
    $p_\theta(e^{s}_{ij} \mid \hat{G}^{s+1}, y) \leftarrow \sum_e q(e^{s}_{ij} \mid \hat{e}^{s+1}_{ij}, e^0_{ij} = e)\, f_\theta(e^0_{ij}=e \mid \hat{G}^{s+1}, y)$\;
    $g_s^X \leftarrow \textsc{PosteriorGumbel}\big(\log p_\theta(X^{s} \mid \hat{G}^{s+1}, y),\; \hat{X}^{s}\big)$ \tcp*[r]{Eqs.~\ref{eq:truncated_gumbel},~\ref{eq:noise_recover} \cite{AStarSampling}}
    $g_s^E \leftarrow \textsc{PosteriorGumbel}\big(\log p_\theta(E^{s} \mid \hat{G}^{s+1}, y),\; \hat{E}^{s}\big)$\;
    $\mathcal{G} \leftarrow \mathcal{G} \cup \{(g_s^X, g_s^E)\}$
}

\vspace{0.15cm}
\tcp{Phase 3: dynamic budget search + guided replay (target condition $y'$)}
\For{$\tau \in \mathcal{T}_{skip}$}{
    $G^\tau \leftarrow \hat{G}^\tau$\;
    \For{$t = \tau$ \KwTo $1$}{
        $p_\theta(x^{t-1}_i \mid G^t, y') \leftarrow \sum_x q(x^{t-1}_i \mid x^t_i, x^0_i = x)\, f_\theta(x^0_i=x \mid G^t, y')$\;
        $p_\theta(e^{t-1}_{ij} \mid G^t, y') \leftarrow \sum_e q(e^{t-1}_{ij} \mid e^t_{ij}, e^0_{ij} = e)\, f_\theta(e^0_{ij}=e \mid G^t, y')$\;
        $X^{t-1} \leftarrow \text{one\_hot}\big(\text{argmax}(\log p_\theta(X^{t-1} \mid G^t, y') + g_{t-1}^X)\big)$\;
        $E^{t-1} \leftarrow \text{one\_hot}\big(\text{argmax}(\log p_\theta(E^{t-1} \mid G^t, y') + g_{t-1}^E)\big)$\;
        $G^{t-1} \leftarrow (X^{t-1}, E^{t-1})$\;
    }
    \If{$f(G^0) == y'$}{
        \Return $G_{CE} \leftarrow G^0$ \tcp*[r]{minimal counterfactual}
    }
}
\Return \text{Failure} \tcp*[r]{no valid counterfactual within budgets}

\caption{Graph Diffusion Counterfactual Explanation via Inversion (GDCE-I)}
\label{alg:method}
\end{algorithm}

\section{Datasets and Classifier Architectures}
\label{app:datasets_classifiers}
We evaluate on four graph-classification datasets, two molecular and two
non-molecular. \textbf{Mutagenicity} \cite{mutagenicity} is a standard molecular
benchmark from the TUDataset
collection~\cite{TUDataset}. \textbf{Benzene}~\cite{BENZENE,BENZENXAI} is a molecular
attribution benchmark whose label indicates the presence of a benzene ring.
\textbf{PROTEINS}~\cite{PROTEINS} and \textbf{TWITTER}
(TWITTER-Real-Graph-Partial)~\cite{TWITTER} are likewise taken from the TUDataset
collection. All four are preprocessed identically for GDCE-I and every baseline, so no method
gains an advantage from a different data representation.

Graphs larger than $50$ nodes are removed. We additionally apply the Huang frequency filter (node types occurring $\le 50$ times globally are dropped, together with the graphs containing them). On Benzene, this removes $36$ graphs containing phosphorus. Mutagenicity and Benzene retain bond types as a $5$-class edge encoding ($0$ = no bond, $1$--$4$ =
single/double/triple/aromatic), whereas PROTEINS and TWITTER carry no edge types and use a binary adjacency. Table~\ref{tab:appendix_datasets} summarizes the resulting statistics.

The two non-molecular datasets follow the same recipe. PROTEINS needs no vocabulary filter. Its nodes carry three secondary-structure labels (helix, sheet, turn), all of them frequent, and the
$50$-node cap leaves $871$ of the $1113$ graphs. TWITTER does need one. Its raw
vocabulary of $1323$ words would make the $(b, n, d_x, d_x)$ attention tensor of the
denoising network exceed device memory at the batch size we use, so we apply the same
frequency rule as for the molecular sets with a threshold of $500$ occurrences. This
leaves $238$ word types and $65031$ graphs, retains $78.9\%$ of all node instances and
leaves the class balance essentially untouched. Nodes are words, edges word
co-occurrences, and the binary label is the sentiment of the tweet, with class $0$
positive and class $1$ negative.

\begin{table}[H]
\centering
\resizebox{\linewidth}{!}{
\begin{tabular}{l | l | c | c | c | c | c}
\toprule
\textbf{Dataset} & \textbf{Source} & \textbf{Graphs (tr/val/test)}
 & \textbf{Node types} & \textbf{Edge classes} & \textbf{Max nodes} & \textbf{Balance (0/1)} \\
\midrule
Mutagenicity & TUDataset~\cite{TUDataset,mutagenicity} & $2377/792/792$ ($3961$) & $13$ & $5$ & $50$ & $57.6/42.4$ \\
Benzene      & \cite{BENZENE,BENZENXAI}                  & $7178/2393/2393$ ($11964$) & $8$ & $5$ & $25$ & $49.9/50.1$ \\
PROTEINS     & TUDataset~\cite{TUDataset,PROTEINS}       & $523/174/174$ ($871$) & $3$ & $2$ & $50$ & $52.9/47.1$ \\
TWITTER      & TUDataset~\cite{TUDataset,TWITTER}        & $39019/13006/13006$ ($65031$) & $238$ & $2$ & $50$ & $52.2/47.8$ \\
\bottomrule
\end{tabular}
}
\caption{Dataset statistics after preprocessing.}
\label{tab:appendix_datasets}
\end{table}

\paragraph{SMILES validity ceiling on Benzene}
Because Benzene molecules are capped at $25$ nodes in the source data, some
aromatic rings are truncated at the boundary. RDKit cannot sanitize a partial
aromatic system, so even a perfectly reconstructed graph may fail to yield a valid
SMILES string. Only $89.05\%$ pass strict sanitization. This is an upper bound on the SMILES column
of Table~\ref{tab:quantitative_results_main} for Benzene. The reported values should be read
against $0.8905$ rather than $1.0$. For comparison, the ceiling on Mutagenicity is
$0.976$.

\paragraph{The shared classifiers}
Every method in Table~\ref{tab:quantitative_results_main} is optimized against, and evaluated by,
the same classifier per dataset. We train two per dataset on the shared training
split: a binary-adjacency \textbf{GCN} (LEConv, $3\times128$ hidden, BatchNorm,
dropout $0.3$, mean pooling), which every method can consume and against which all
of them are measured, and an edge-feature-aware \textbf{GINE} (GINEConv with a
$5$-class edge encoder, otherwise identical), reserved for the two methods that
predict bond types (GDCE-I and UCExplainer). PROTEINS and TWITTER carry no edge types and
therefore have a GCN only. Both are plain PyTorch modules with no framework
dependency, so the identical weights are loaded inside each baseline's own
container. We verified that all five repositories reproduce the same predictions on
the same graphs before running any evaluation. Test accuracies: Mutagenicity
$0.806$ (GCN) / $0.813$ (GINE), Benzene $0.919$ (GCN) / $1.000$ (GINE),
PROTEINS $0.632$ (GCN), TWITTER $0.648$ (GCN).

Because every method is trained and scored against these classifiers, class labels
are taken from the classifier rather than from the dataset annotation, so that
``flipping the label'' and ``flipping the classifier'' coincide by construction.

\end{document}